\documentclass[runningheads]{llncs}
\usepackage{graphicx}
\usepackage{multirow}
\usepackage{tikz}
\usepackage{comment}
\usepackage{amsmath,amssymb} 
\usepackage{bbm}
\usepackage{color}

\usepackage[accsupp]{axessibility}  

\PassOptionsToPackage{hyphens}{url}\usepackage{hyperref}

\begin{document}
\pagestyle{headings}
\mainmatter
\def\ECCVSubNumber{100}  

\title{Where a Strong Backbone Meets Strong Features -- ActionFormer for Ego4D Moment Queries Challenge} 

\titlerunning{ActionFormer for Ego4D Moment Queries} 
\authorrunning{ECCV-22 submission ID \ECCVSubNumber} 
\author{Anonymous ECCV submission}
\institute{Paper ID \ECCVSubNumber}


\titlerunning{ActionFormer for Ego4D Moment Queries}
%
\author{Fangzhou Mu \and
Sicheng Mo \and
Gillian Wang \and
Yin Li}
\authorrunning{Mu et al.}
%
\institute{University of Wisconsin-Madison}

\maketitle

\begin{abstract}
This report describes our submission to the Ego4D Moment Queries Challenge 2022. Our submission builds on ActionFormer~\cite{zhang2022actionformer}, the state-of-the-art backbone for temporal action localization, and a trio of strong video features from SlowFast~\cite{Feichtenhofer_2019_ICCV}, Omnivore~\cite{girdhar2022omnivore} and EgoVLP~\cite{kevin2022egovlp}. Our solution is ranked 2$nd$ on the public leaderboard with 21.76\% average mAP on the test set, which is nearly three times higher than the official baseline. Further, we obtain 42.54\% Recall@1x at tIoU=0.5 on the test set, outperforming the top-ranked solution by a significant margin of 1.41 absolute percentage points. Our code is available at \url{ https://github.com/happyharrycn/actionformer\_release}.

\end{abstract}

\section{Introduction}

Given an untrimmed egocentric video, the Ego4D Moment Queries (MQ) task seeks to localize all moments of actions in time and recognize their categories. Known as temporal action localization in third-person video understanding, this challenging task is often approached in two steps owing to the prohibitive cost of end-to-end training. Clip-level features are first extracted from raw video frames using a pre-trained feature network such as SlowFast~\cite{Feichtenhofer_2019_ICCV}. A dedicated temporal network subsequently makes sense of the one-dimensional feature sequence and predicts the onset and offset of action instances as well as their categories.

Following the same two-step approach, our solution zooms in on its two key components, namely \textit{video features} and \textit{temporal network backbone}. Until recently, feature networks~\cite{Feichtenhofer_2019_ICCV,girdhar2022omnivore,arnab2021vivit,fan2021multiscale} were predominantly trained on third-person video datasets such as Kinetics~\cite{kay2017kinetics}. Their learned motion cues and action statistics are hence not fully representative of first-person videos, presenting major challenge to egocentric video understanding. This gap has recently been closed by EgoVLP~\cite{kevin2022egovlp}, a dedicated egocentric pre-training method, following the release of the Ego4D dataset~\cite{Ego4D2022CVPR}. Meanwhile, our prior work of ActionFormer~\cite{zhang2022actionformer}, a transformer-based backbone, recently established the state-of-the-art results for temporal action localization. ActionFormer adopts local self-attention for temporal reasoning and captures actions of variable length using a flexible point-based action representation. Thanks to this orchestrated design, ActionFormer performs exceptionally well on the egocentric EPIC-Kitchens dataset~\cite{Damen2022RESCALING} and serves as the backbone for many winning entries in the EPIC-Kitchens 100 Action Detection Challenge 2022~\footnote{\url{https://epic-kitchens.github.io/2022}}. 

To this end, we take a natural step to bring together the best of both worlds. Our submission to the Ego4D MQ Challenge is a straightforward extension of ActionFormer with strong video features from both third-person and first-person video pre-training. Our submission is ranked 2$nd$ on the public leaderboard with 21.76\% average mAP on the test set, which is nearly three times higher than the official baseline. Further, we obtain 42.54\% Recall@1x at tIoU=0.5 on the test set, outperforming the top-ranked solution by a significant margin of 1.41 absolute percentage points.

\section{Approach}

\begin{figure*}[t!]
	\centering 
	\includegraphics[width=0.8\linewidth]{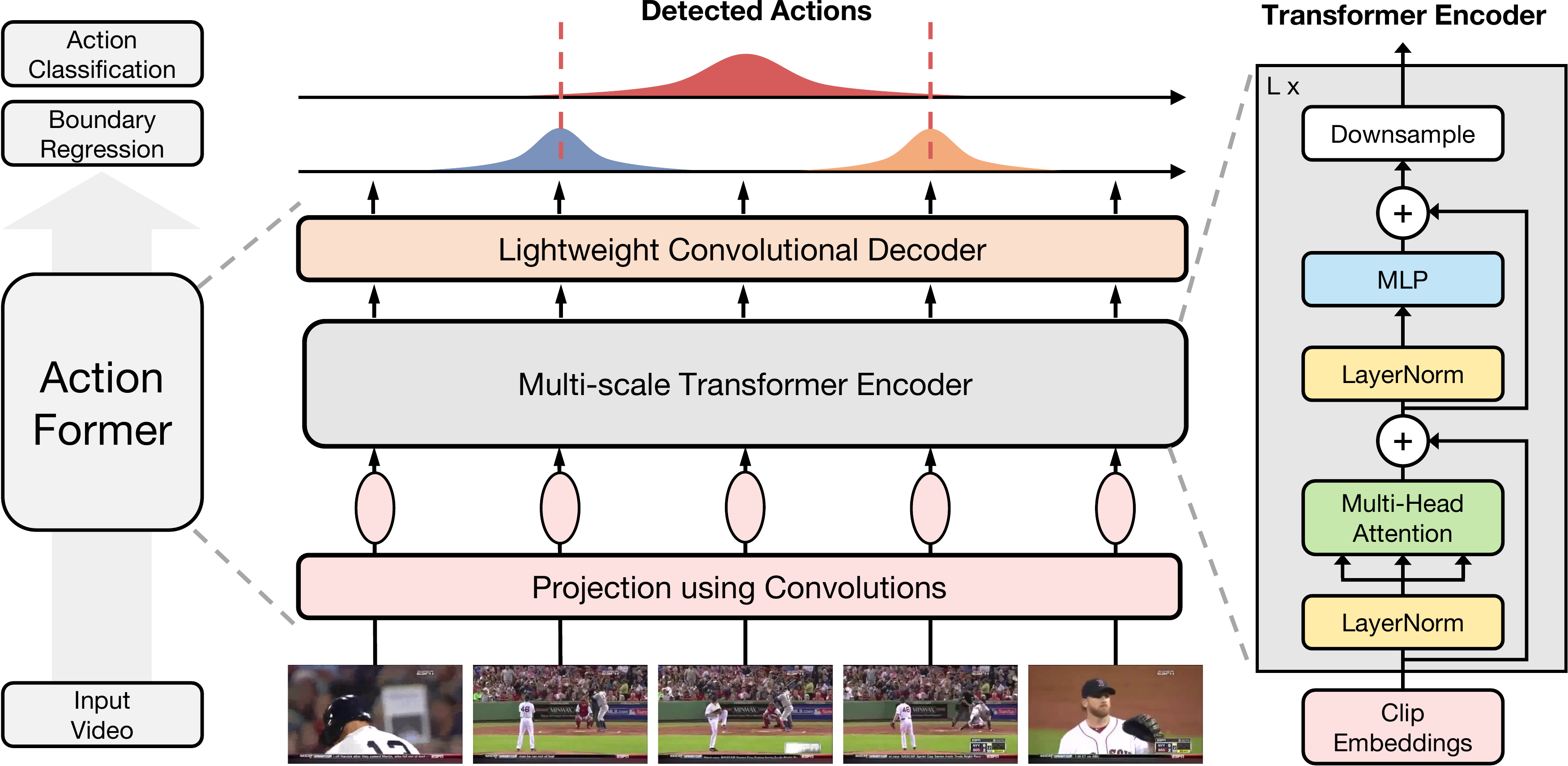}\vspace{-0.5em}
	\caption{Overview of ActionFormer. Taken from~\cite{zhang2022actionformer}.}
	\label{fig:overview}\vspace{-1.5em}
\end{figure*}

Our solution first extracts clip-level video features using pre-trained networks. A video is thus represented as a sequence of feature vectors, where each feature vector is a compact representation of a multi-frame video clip. ActionFormer, as shown in Figure~\ref{fig:overview}, subsequently takes this sequence as input, builds a multi-scale feature pyramid using local self-attentions and interprets every point on the pyramid as an action candidate. The candidate points are classified as either one of the pre-defined action categories or the background. The distances from foreground points to the action's onset and offset are further regressed.\smallskip

\noindent \textbf{Feature Extraction.} We leverage three strong feature networks for clip-level feature extraction, including SlowFast with ResNet-101 backbone~\cite{Feichtenhofer_2019_ICCV} and Omnivore with Swin-L backbone~\cite{girdhar2022omnivore} pre-trained on the third-person videos from Kinetics-400~\cite{kay2017kinetics}, and the video encoder of EgoVLP~\cite{kevin2022egovlp} pre-trained on Ego4D~\cite{Ego4D2022CVPR}. We use the official release of SlowFast and Omnivore features~\footnote{\url{https://ego4d-data.org/docs/data/features/}} without fine-tuning, and the pre-trained weights from the official EgoVLP code repository~\footnote{\url{https://github.com/showlab/EgoVLP}} for feature extraction. On 30 FPS videos, SlowFast and Omnivore features are extracted from clips of 32 frames with a stride of 16. EgoVLP features are extracted from non-overlapping clips of 4 frames following the official implementation. The feature dimensions for SlowFast, Omnivore and EgoVLP are 2304, 1536 and 256, respectively.\smallskip

\noindent \textbf{Feature Fusion.} Our model learns a set of linear projections to reduce feature dimension before feeding them into ActionFormer. Specifically, SlowFast, Omnivore and EgoVLP features are \textit{independently} projected to 386, 386 and 256 dimensions and then concatenated together as the input to ActionFormer. We found this works better than na\"ively stacking the features before projection.\smallskip

\noindent \textbf{Action Localization with ActionFormer.} ActionFormer first embeds the projected clip-level features with learned convolutions. The embedded features are then encoded into a multi-scale feature pyramid using local self-attentions, yielding a hierarchy of candidate action moments with varying time spans. The candidate moments are further examined by convolutional heads shared across pyramid levels for action classification and boundary regression, from which action segments are assembled and combined using multi-class SoftNMS~\cite{Bodla_2017_ICCV}. We refer readers to the main paper~\cite{zhang2022actionformer} for more technical details.\smallskip

\noindent \textbf{Implementation Details.} Our instantiation of ActionFormer uses a 6-level feature pyramid where the finest temporal resolution is 1,024 time steps. The embedding dimension is 1,024 throughout the model and the transformer layers use 16 attention heads. For our final submission, we train ActionFormer on the combined training and validation data for 15 epochs with the AdamW optimizer~\cite{loshchilov2017decoupled} using a mini-batch size of 2 and a learning rate of 1e-4 with linear warm-up and cosine annealing. The number of predicted segments for each video is capped at 2,000. We report test results using the checkpoint at epoch 10 based on earlier validation results.

\begin{table}[t]
\centering
\caption{\textbf{Results on the Ego4D MQ task.} Our model outperforms the baseline~\cite{zhao2021video} by a wide margin with the same SlowFast features. Further performance gain comes from combining features from multiple sources and improving feature fusion. `Cat' stands for concatenation, and `Proj' for independent linear projection of source features.}
\resizebox{0.95\textwidth}{!}
{
\begin{tabular}{c|l|c|c|c} 
\hline
\textbf{Split}        & \multicolumn{1}{c|}{\textbf{Method \& Features}} & \textbf{Fusion}      & \begin{tabular}[c]{@{}c@{}}\textbf{Average}\\\textbf{mAP}\end{tabular} & \begin{tabular}[c]{@{}c@{}}\textbf{Recall@1x}\\\textbf{(tIoU=0.5)}\end{tabular}  \\ 
\hline
\multirow{5}{*}{Val}  & VSGN~\cite{zhao2021video} w.~SlowFast                              & \multirow{2}{*}{n/a} & 6.03                                                                   & 25.16   \\
& Ours~\cite{zhang2022actionformer} w.~SlowFast                              &                      & 13.24                                                                  & 27.80 \tabularnewline \cline{2-5}                                                                            
& Ours~\cite{zhang2022actionformer} w.~SlowFast+Omnivore                     & \multirow{2}{*}{Cat} & 17.17                                                                  & 33.46                                                                            \\
& Ours~\cite{zhang2022actionformer} w.~SlowFast+Omnivore+EgoVLP~               &                      & 20.90                                                                  & 36.84 \tabularnewline \cline{2-5}                                                                             
& Ours~\cite{zhang2022actionformer} w.~SlowFast+Omnivore+EgoVLP~               & ~Proj+Cat~            & \textbf{21.40}                                                         & \textbf{38.73}                                                                   \\ 
\hline\hline
\multirow{3}{*}{Test} & VSGN~\cite{zhao2021video} w.~SlowFast                              & n/a                  & 5.68                                                                   & 24.25 \tabularnewline \cline{2-5}   
& Ours~\cite{zhang2022actionformer} w.~SlowFast+Omnivore                     & Cat                  & 17.92                                                                  & 35.51 \tabularnewline\cline{2-5}   
& Ours~\cite{zhang2022actionformer} w.~SlowFast+Omnivore+EgoVLP~               & ~Proj+Cat~            & \textbf{21.76}                                                         & \textbf{42.54} \\
\hline
\end{tabular}\label{tab:results}
}\vspace{-1.5em}
\end{table}

\section{Experiments and Results}

We now present our experiments and results for the Ego4D MQ Challenge.\smallskip

\noindent \textbf{Dataset.} The Ego4D MQ dataset includes a total 326.4 hours of videos, 2,488 video clips and 22.2K action instances from 110 pre-defined action categories. The ratio of video clips is 6:2:2 among the train/val/test splits. The action duration follows a long-tail distribution with a majority of instances less than one minute and 22.4\% shorter than 3 seconds.\smallskip

\noindent \textbf{Evaluation Protocol and Metrics.} We follow the official train/val/test splits for evaluation. We train on the train split for evaluation on the val split, and train on the combined train and val splits for evaluation on the test split. We calculate mean average precision (mAP) at tIoU thresholds [0.1:0.1:0.5] and report their average, i.e., the average mAP. We further report Recall@1x at tIoU=0.5, where x stands for the number of ground-truth instances for an action category in one video. This metric measures the percentage of ground-truth instances that have at least one prediction with tIoU greater than 0.5 within the Top-x results for the target action category.\smallskip

\noindent \textbf{Results.} Table~\ref{tab:results} summarizes our results on the val and test splits. We observe substantial improvement in both metrics as we bundle more features together. In particular, the addition of EgoVLP features brings a large performance boost (3.76\% in average mAP and 7.03\% in Recall) on the test split. This validates the effectiveness of egocentric video pre-training on downstream tasks. Further, our results demonstrate that feature fusion matters, and separate projection of features yields better results. We hypothesize that features from difference sources are better balanced with explicit control of the projected feature dimension. Our final solution reaches an average mAP of 21.76\% which almost quadruples the official baseline result of 5.68\%. In addition, we arrive at 42.54\% Recall@1x at tIoU=0.5, almost doubling the baseline result of 24.25\%. Compared to the top-ranked submission, our solution is 1.83 absolute percentage points lower in average mAP but 1.41 points higher in Recall@1x at tIoU=0.5. Encouraged by our observation, we conjecture that the top-ranked submission makes use of stronger features.\smallskip

\section{Conclusion}

In this report, we described our solution using ActionFormer and strong video features for the Ego4D MQ task. Our solution considers a latest temporal action localization model in tandem with new video features, resulting in a simple model with strong empirical results. We hope that our solution and results can provide insight into the Ego4D MQ task.\smallskip 

\noindent \textbf{Limitations and Discussion}. Our solution approaches the MQ task as a special case of temporal action localization, treating each activity as a separate category and thus ignoring their text descriptions and relationships. We conjecture that further integration of a language model might better contextualize these activities. Further, our method does not explicitly model visual cues unique in egocentric videos, such as ego-motion, egocentric attention, hand trajectory, or object presence, which have proven effective for egocentric vision~\cite{li2021eye}.

%
%
\bibliographystyle{splncs04}
\bibliography{egbib}

\begin{thebibliography}{10}
\providecommand{\url}[1]{\texttt{#1}}
\providecommand{\urlprefix}{URL }
\providecommand{\doi}[1]{https://doi.org/#1}

\bibitem{arnab2021vivit}
Arnab, A., Dehghani, M., Heigold, G., Sun, C., Lu{\v{c}}i{\'c}, M., Schmid, C.:
  Vivit: A video vision transformer. In: International Conference on Computer
  Vision (ICCV) (2021)

\bibitem{Bodla_2017_ICCV}
Bodla, N., Singh, B., Chellappa, R., Davis, L.S.: Soft-nms -- improving object
  detection with one line of code. In: International Conference on Computer
  Vision (ICCV) (2017)

\bibitem{Damen2022RESCALING}
Damen, D., Doughty, H., Farinella, G.M., , Furnari, A., Ma, J., Kazakos, E.,
  Moltisanti, D., Munro, J., Perrett, T., Price, W., Wray, M.: Rescaling
  egocentric vision: Collection, pipeline and challenges for epic-kitchens-100.
  International Journal of Computer Vision (IJCV)  (2022)

\bibitem{fan2021multiscale}
Fan, H., Xiong, B., Mangalam, K., Li, Y., Yan, Z., Malik, J., Feichtenhofer,
  C.: Multiscale vision transformers. In: International Conference on Computer
  Vision (ICCV) (2021)

\bibitem{Feichtenhofer_2019_ICCV}
Feichtenhofer, C., Fan, H., Malik, J., He, K.: Slowfast networks for video
  recognition. In: International Conference on Computer Vision (ICCV) (2019)

\bibitem{girdhar2022omnivore}
Girdhar, R., Singh, M., Ravi, N., van~der Maaten, L., Joulin, A., Misra, I.:
  {Omnivore: A Single Model for Many Visual Modalities}. In: Computer Vision
  and Pattern Recognition (CVPR) (2022)

\bibitem{Ego4D2022CVPR}
Grauman, K., Westbury, A., Byrne, E., Chavis, Z., Furnari, A., Girdhar, R.,
  Hamburger, J., Jiang, H., Liu, M., Liu, X., Martin, M., Nagarajan, T.,
  Radosavovic, I., Ramakrishnan, S.K., Ryan, F., Sharma, J., Wray, M., Xu, M.,
  Xu, E.Z., Zhao, C., Bansal, S., Batra, D., Cartillier, V., Crane, S., Do, T.,
  Doulaty, M., Erapalli, A., Feichtenhofer, C., Fragomeni, A., Fu, Q., Fuegen,
  C., Gebreselasie, A., Gonzalez, C., Hillis, J., Huang, X., Huang, Y., Jia,
  W., Khoo, W., Kolar, J., Kottur, S., Kumar, A., Landini, F., Li, C., Li, Y.,
  Li, Z., Mangalam, K., Modhugu, R., Munro, J., Murrell, T., Nishiyasu, T.,
  Price, W., Puentes, P.R., Ramazanova, M., Sari, L., Somasundaram, K.,
  Southerland, A., Sugano, Y., Tao, R., Vo, M., Wang, Y., Wu, X., Yagi, T.,
  Zhu, Y., Arbelaez, P., Crandall, D., Damen, D., Farinella, G.M., Ghanem, B.,
  Ithapu, V.K., Jawahar, C.V., Joo, H., Kitani, K., Li, H., Newcombe, R.,
  Oliva, A., Park, H.S., Rehg, J.M., Sato, Y., Shi, J., Shou, M.Z., Torralba,
  A., Torresani, L., Yan, M., Malik, J.: Ego4d: Around the world in 3,000 hours
  of egocentric video. In: Computer Vision and Pattern Recognition (CVPR)
  (2022)

\bibitem{kay2017kinetics}
Kay, W., Carreira, J., Simonyan, K., Zhang, B., Hillier, C., Vijayanarasimhan,
  S., Viola, F., Green, T., Back, T., Natsev, P., et~al.: The kinetics human
  action video dataset. arXiv preprint arXiv:1705.06950  (2017)

\bibitem{li2021eye}
Li, Y., Liu, M., Rehg, J.: In the eye of the beholder: Gaze and actions in
  first person video. Transactions on Pattern Analysis and Machine Intelligence
  (TPAMI)  (2021)

\bibitem{kevin2022egovlp}
Lin, K.Q., Wang, A.J., Soldan, M., Wray, M., Yan, R., Xu, E.Z., Gao, D., Tu,
  R., Zhao, W., Kong, W., Cai, C., Wang, H., Damen, D., Ghanem, B., Liu, W.,
  Shou, M.Z.: Egocentric video-language pretraining. In: Neural Information
  Processing Systems (NeurIPS) (2022)

\bibitem{loshchilov2017decoupled}
Loshchilov, I., Hutter, F.: Decoupled weight decay regularization. In:
  International Conference on Learning Representations (ICLR) (2019)

\bibitem{zhang2022actionformer}
Zhang, C., Wu, J., Li, Y.: Actionformer: Localizing moments of actions with
  transformers. In: European Conference on Computer Vision (ECCV) (2022)

\bibitem{zhao2021video}
Zhao, C., Thabet, A.K., Ghanem, B.: Video self-stitching graph network for
  temporal action localization. In: International Conference on Computer Vision
  (ICCV) (2021)

\end{thebibliography}
\end{document}